\documentclass[journal]{IEEEtran}
\usepackage{cite}

\usepackage{amssymb,amsfonts}
\usepackage{algorithmic}
\usepackage{graphicx}
\usepackage{textcomp}
\usepackage{import}
\usepackage[super]{nth}
\usepackage{xcolor}
\usepackage{booktabs}
\usepackage{multirow}
\usepackage{pifont}
\usepackage{enumerate}
\usepackage{amsmath}
\usepackage{paralist}
\usepackage{subcaption}

\DeclareMathOperator*{\argmin}{argmin}
\DeclareMathOperator*{\argmax}{argmax}
\usepackage{marvosym}
\usepackage{enumitem}

\begin{document}
\title{COIN: Contrastive Identifier Network for Breast Mass Diagnosis in Mammography}
\author{
Heyi Li, Dongdong Chen, William H. Nailon,  Mike E. Davies \IEEEmembership{Fellow, IEEE}, and David Laurenson 
\thanks{H. Li, D. Chen, William H. Nailon, Mike E. Davies, and David Laurenson are with the School of Engineering, the University of Edinburgh University, Edinburgh, EH9 3JL, U.K. (e-mail: \{Heyi. Li, D.Chen, W.Nailon, Mike.Davies, Dave.Laurenson\}@ed.ac.uk).}
\thanks{This work was supported in part by the ERC C-SENSE project ERCADG-2015-694888. }}

\maketitle

\begin{abstract}
Computer-aided breast cancer diagnosis in mammography is a challenging problem, stemming from mammographical data scarcity and data entanglement. 
In particular, data scarcity is attributed to the privacy and expensive annotation. And data entanglement is due to the high similarity between benign and malignant masses, of which manifolds reside in lower dimensional space with very small margin. 
To address these two challenges, we propose a deep learning framework, named Contrastive Identifier Network  (\textsc{COIN}), which integrates adversarial augmentation and manifold-based contrastive learning. 
Firstly, we employ adversarial learning to create both on- and off-distribution mass contained ROIs. 
After that, we propose a novel contrastive loss with a built Signed graph. 
Finally, the neural network is optimized in a contrastive learning manner, with the purpose of improving the deep model's discriminativity on the extended dataset.
In particular, by employing COIN, data samples from the same category are pulled close whereas those with different labels are pushed further in the deep latent space. 
Moreover, COIN outperforms the state-of-the-art related algorithms for solving breast cancer diagnosis problem by a considerable margin, achieving 93.4\% accuracy and 95.0\% AUC score.  The code will release on ***. 
\end{abstract}


\section{Introduction}
Breast cancer is widely acknowledged as the most frequently diagnosed cancer and the second fatal disease for women around the world \cite{IARC2008}. 
Although no effective method has been discovered for prevention, mammography screening is advantageous to early breast mass diagnosis (BMD), which has practically increased the associated survival rates along with early treatments \cite{oliver2010review}. 
Screening mammography is particularly useful when tumours are invasive (measuring $<2$ cm) and too small to be palpable or cause symptoms \cite{breaststa2014}. 
However, manual interpretations have been limited by wide variations in pathology and the potential fatigue of human experts \cite{oliver2010review}.
Double reading is thereby employed in many western countries \cite{blanks1998comparison, brown1996mammography}, which has been proven to increase both sensitivity and specificity for the interpretations. 
In recent years, computer-assisted interventions have been designed and employed to benefit researchers and doctors as an alternative to a human double reader for an optimal healthcare \cite{shen2017deep,mckinney2020international}. 

\begin{figure}
    \centering
    \includegraphics[width=0.45\textwidth]{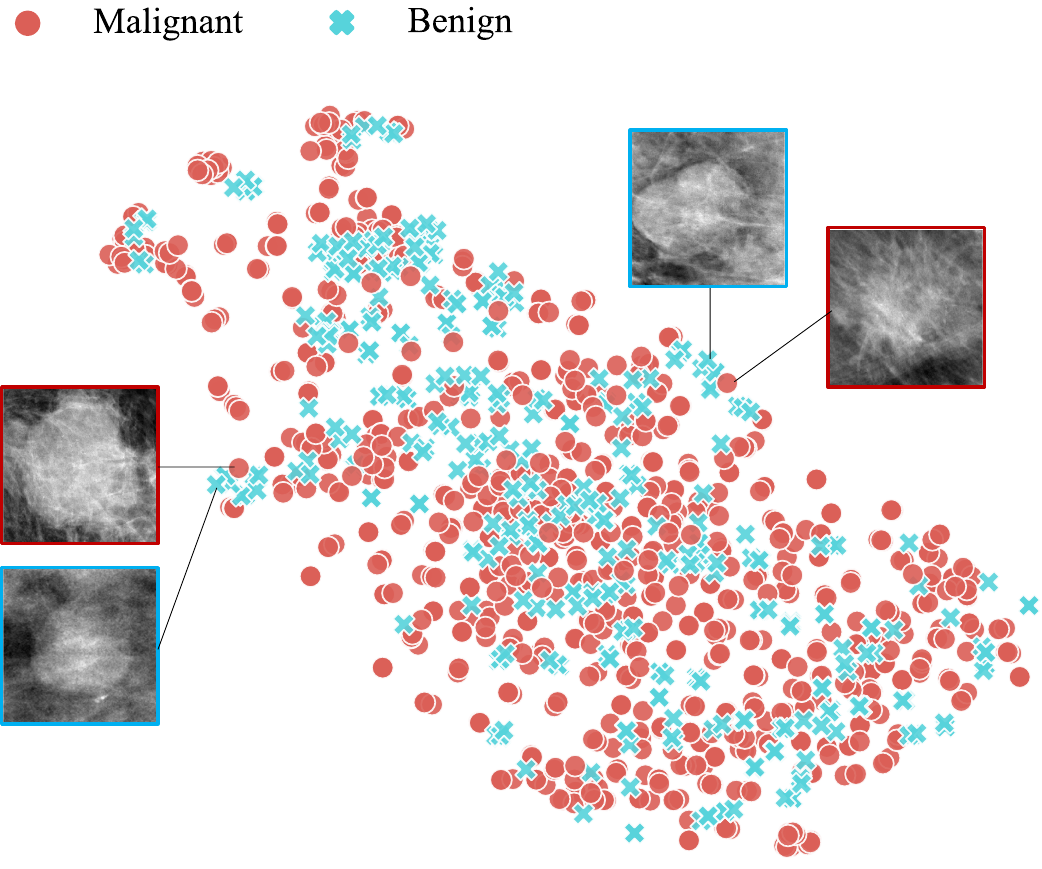}
    \caption{The illustration of BMD challenges with an INbreast dataset example: \textit{$Q_1$ -  data scarcity}  and \textit{$Q_2$ - data entanglement}. Red stands for the 2D t-SNE \cite{maaten2008visualizing} embedding of malignant masses and blue for that of benign lesions. The four images are corresponding mass examples. }
    \label{fig:embed_originaldata}
\end{figure}

\subsection{Classical Methods for BMD} 
Breast mass classification between benign and malignant lesions is one of the most important and challenging tasks for commercial computer aided diagnosis systems (CADs). This is not only because of the small proportion of cancerous cases among all screenings, but also due to their high similarities. 
This characteristic can be illustrated as Fig. \ref{fig:embed_originaldata}, where benign and malignant masses are visually very similar as well as they embed in an intersected manner with t-SNE visualization \cite{maaten2008visualizing}.
Although the speed of development in CADs has not been as rapid as that of medical imaging techniques, the situation has improved as machine learning approaches advancing \cite{jiao2018parasitic}. 
When dealing with the classification or diagnosis task,
finding or learning distinctive features of cancerous masses and their surrounding tissues is the most important task, so that inherent regularities or patterns can be well described \cite{oliver2010review}. 
Traditionally, meaningful features were hand engineered by domain experts \cite{varela2006use}, which instill task-specific knowledge \cite{kooi2017large}.
However, the major cons of this process is clear that engineers of machine learning have to exploit essential algorithms with the help from medical domain experts.   
Additionally, manual designed features  may lead to strong bias for the training of the algorithm, resulting in limited performance \cite{jalalian2013computer}, \emph{e.g.} high false positive rate and low specificity \cite{malich2006cad}.

\subsection{Deep Learning Methods for BMD} 
In recent years, owing to the success of deep neural networks (deep learning) \cite{lecun2015deep} applied in various computer perception tasks \cite{chen2017unsupervised}, a noticeable shift from rule-based, problem specific solutions to increasingly generic, problem agnostic-based algorithms  has been seen in mammographical CADs \cite{carneiro2017automated, shams2018deep, zhu2017deep, arevalo2016representation, kooi2017discriminating, lotter2017multi,chen2020decomposition,chen2020compressive}. 
Specifically, \cite{arevalo2016representation} and \cite{kooi2017discriminating} claimed that features extracted by a CNN can achieve better performance for breast mass discrimination, when compared to various hand-crafted features.
However, passing through the bottleneck in lower dimension of classifying a mammographical mass is very difficult in CNN models, yielding imprecise predictions.
This is not only because of the low signal-to-noise ratio of the screening images like other medical imaging modalities \cite{oliver2010review}, but breast masses in mammography are also suffered from two other major problems: 
\begin{itemize}
    \item \textbf{\emph{$Q_1$ - Data Scarcity}} 
    \cite{li2018improved, dhungel2016automated}, which is difficult to solve due to the issue of patients' privacy and the tremendous workloads of annotation by human experts; 
    
    \item \textbf{\emph{$Q_2$ - Data entanglement}}. 
    It is very challenging when compares to natural image recolonization problems, which is attributed to the small margin between benign and malignant data manifolds (Fig. \ref{fig:embed_originaldata}).  
    
\end{itemize}
The detailed recent efforts that have been made on these two major problems will be discussed in Sec. \ref{Sec:RelatedWork}.

\subsection{Our contribution}
Based on all of the above observations, in this paper, we propose a new deep convolutional neural network, called  \emph{Contrastive Identifier Network} (\textsc{COIN}), in which the contrastive learning and manifold learning are integrated for breast mass classification (benign vs. malignant).
In particular, we propose to employ the adversarial learning for data augmentation, so that both on- and off-manifold new samples with more distinctive features are created in an unsupervised fashion;
We propose a novel triplet contrastive loss, which exploits the merit of the Signed similarity graph. In such a way, the locality of the manifold is approximated as the built deep network being trained. 
By incorporating these two methods into the deep neural network, we solve the manifold embedding problem by a learning process, instead of computing the expensive eigenvalue decomposition for standard graph spectral learning \cite{chen2014local}.
By integrating these two methods, features discriminativity is improved in deep latent space (Fig. \ref{fig:diagnet_motivations}). 
In particular, data samples from the same class are pulled close, meanwhile those with different labels are pushed away in the deep latent space. 
Consequently, the intra-class difference is minimized, and more importantly, the inter-class manifold margin is maximized in the deep representation space.
A preliminary version of this work appeared in \cite{li2019signed}.
This paper extends \cite{li2019signed} by discussion and experiments so as to prove the effectiveness of our motivation for solving data scarcity (Q1) and data entanglement (Q2).

\section{Related Work}
\label{Sec:RelatedWork}
In this section, we will introduce the existing solutions and their limitations for the purpose of solving $Q_1$-Data Scarcity and $Q_2$-Data Entanglement.  
\subsection{Approaches to $Q_1$} In order to alleviate the data scarcity problem,  works in 
\cite{dhungel2016automated, zhu2017deep, li2019deep, li2018improved} have applied classical affine or elastic transformations for data augmentation in mammography (e.g. flips, rotations, random crops, etc.). These methods are straightforward and effective for increasing the total amount of training data.  However, the distributions of the generated samples are not clear.  Generated samples from unknown distributions are likely to cause an even worse generalization \cite{wong2016understanding}. 
Accordingly,  adversarial learning \cite{goodfellow2014generative} has been employed to generate synthetic images on the manifold of real mammograms, benefiting from the powerful ability to learn the underlying distribution implicitly without modeling the original data prior. 
So far, there is only one application on mammography has been noticed to automatically solve the breast mass classification problem \cite{wu2018conditional}, in which both benign and malignant mass-contained ROIs are created by a conditional generative adversarial net (GAN). 
However, the performance is less encouraging. Their experiments have shown a  limited AUC score improvement, when compared to conventional augmentation methods \cite{wu2018conditional}.
This is potentially because GAN-based augmentations disregard the importance of off-distribution samples, that locate closely to the real data manifold \cite{yu2017open}. We believe these off-distribution samples may also play a very important role in increasing discriminativity while training the model.

\subsection{Approaches to $Q_2$} In order to mitigate the challenge of data entanglement,  many efforts have been tried with CNNs for increasing the discriminativity of latent features in BCD prlblem. 
For example, some researchers have proposed the use of extracting segmentation-related features by CNNs, either with radiologists' pixel-level annotations \cite{dhungel2016automated} or with the generated semantic masks from automatic segmentation algorithms \cite{li2019deep}.
This type of algorithms was originally inspired by the essential of shape and boundary hand-crafted features \cite{oliver2010review}. 
Although these algorithms have improved diagnosis performance, they are typically complicated to construct, either due to their multiple-problem structures, multiple-phase training or large number of parameters. And these  are especially challenging for medical experts. 
More recently, contrastive learning has shown great promise as a type of powerful discriminative approach in various types of computer vision models \cite{xie2020self, hjelm2018learning, henaff2019data, he2020momentum, khosla2020supervised}.
Nevertheless, this method has never been employed in any mammography-related problems as far as we acknowledge. 
In essence, the family of contrastive objective functions aim to enlarge the distances of feature vector pairs in the deep latent space by a self-supervised manner \cite{he2020momentum}. 
Although feature vectors can be separated apart from each other by this technique, inherent  structural and geometrical features of data are ignored, thus features in latent space cannot be enhanced across various classes.
Manifold learning, on the other hand, can mitigate this dilemma by preserving the data topological locality \cite{chen2017unsupervised}. It is widely employed as a non-linearly dimensionality reduction method, since data typically resides on a low-dimensional manifold embedded into a high-dimensional ambient space in real applications \cite{seung2000manifold}.
However, there are few approaches using manifold learning to solve classification problem. 
In fact, there are neither studies on manifold analysis for mammography nor using manifold learning to alleviate the high data similarity problem. 
Thereby, it is very meaningful to do some preliminary studies on using manifold learning for mammography screening diagnosis.

\section{Methodology}
\begin{figure*}[htb!]
    \centering
    \includegraphics[width=0.98\textwidth]{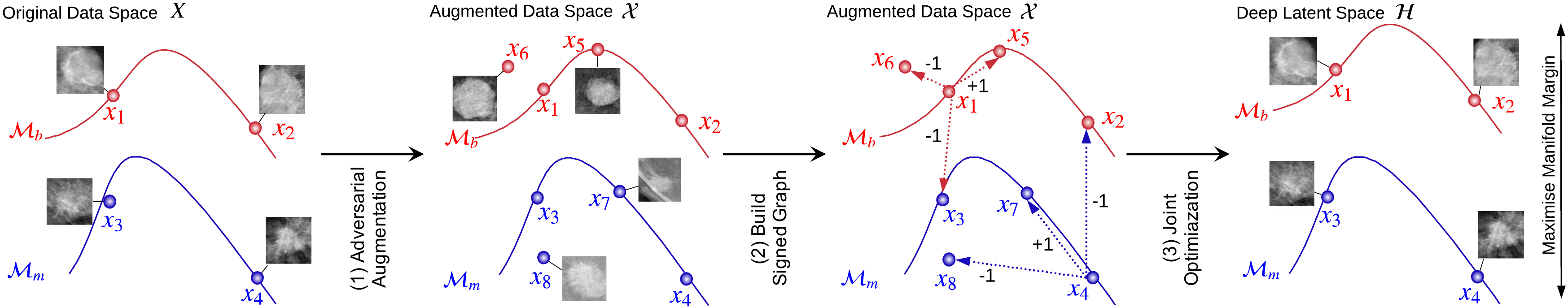}
    \caption{An illustration  of the proposed \textsc{COIN} framework for BMD, which consists of three steps: adversarial augmentation, to build Signed graph, and the joint optimization. 
    In the figure $\{x_1, x_2\}$  are samples on benign manifold $\mathcal{M}_b$ and $\{x_3, x_4\}$ are on malignant manifold $\mathcal{M}_m$. 
    In the first step (adversarial data augmentation), positive neighbors $x_5$ and $x_7$ are created  with  Eq. (\ref{eq:positiveneighbors_A}) for benign and malignant manifold, separately; and negative neighbors $x_6$ and $x_8$ are generated with either Eq. (\ref{eq:negativeneighbors_A}) for benign and malignant manifold, individually.  After that, a signed graph is built upon both original and augmented samples as Eq. (\ref{eq:diagnet_adjacency}). Finally, the joint loss as Eq. (\ref{eq:signedlap_loss}) is optimized in the deep latent space, so that the margin between benign and malignant manifold are maximized. }
    \label{fig:diagnet_motivations}
\end{figure*}

In this section, after discussing the notations and problem formulation utilized in this paper, we formally introduce the details of \textsc{COIN}, which consists of three steps as demonstrated in Fig. \ref{fig:diagnet_motivations}):
\begin{inparaenum}[1)]
\item adversarial augmentation for mammography,
\item a signed graph Laplacian built upon the augmented data, 
\item the proposed contrastive loss and the overall objective function.
\end{inparaenum}
Additionally, we also present the details of constructing the deep network and corresponding implementation. 

\subsection{Notations and Problem Formulation}
Given a dataset $\mathcal{D}=\{(\mathbf{x}_i, \mathrm{y}_i)\}_{i=1}^{N}$, $\mathbf{x}_i \in \mathbb{R}^{H\times W}$ is a real-valued grayscale ROI, and $\mathrm{y}_i$ is the corresponding mass diagnosis label. Note that each ROI  contains only one mass cropped and resized into the fixed size $H\times W$ from a certain mammogram, where $H$ and $W$  both equal to $224$.
With the defined dataset $\mathcal{D}$, let $D_c = \{(\mathbf{x}_{i,c}, \mathrm{y}_{i,c})\}_{i=1}^{N_c}$ be the sub-dataset with $N_c$ samples from the $c$-th category, where $c \in \{0: \textrm{Benign},1:\textrm{Malignant}\}$, and $\mathbf{x}_{i,c} \in X_c$ and $\mathrm{y}_{i,c} \in Y_c$ are arbitrary data sample and its label in this sub-dataset.

\begin{figure}[!t]
\centering
	\includegraphics[width=0.48\textwidth]{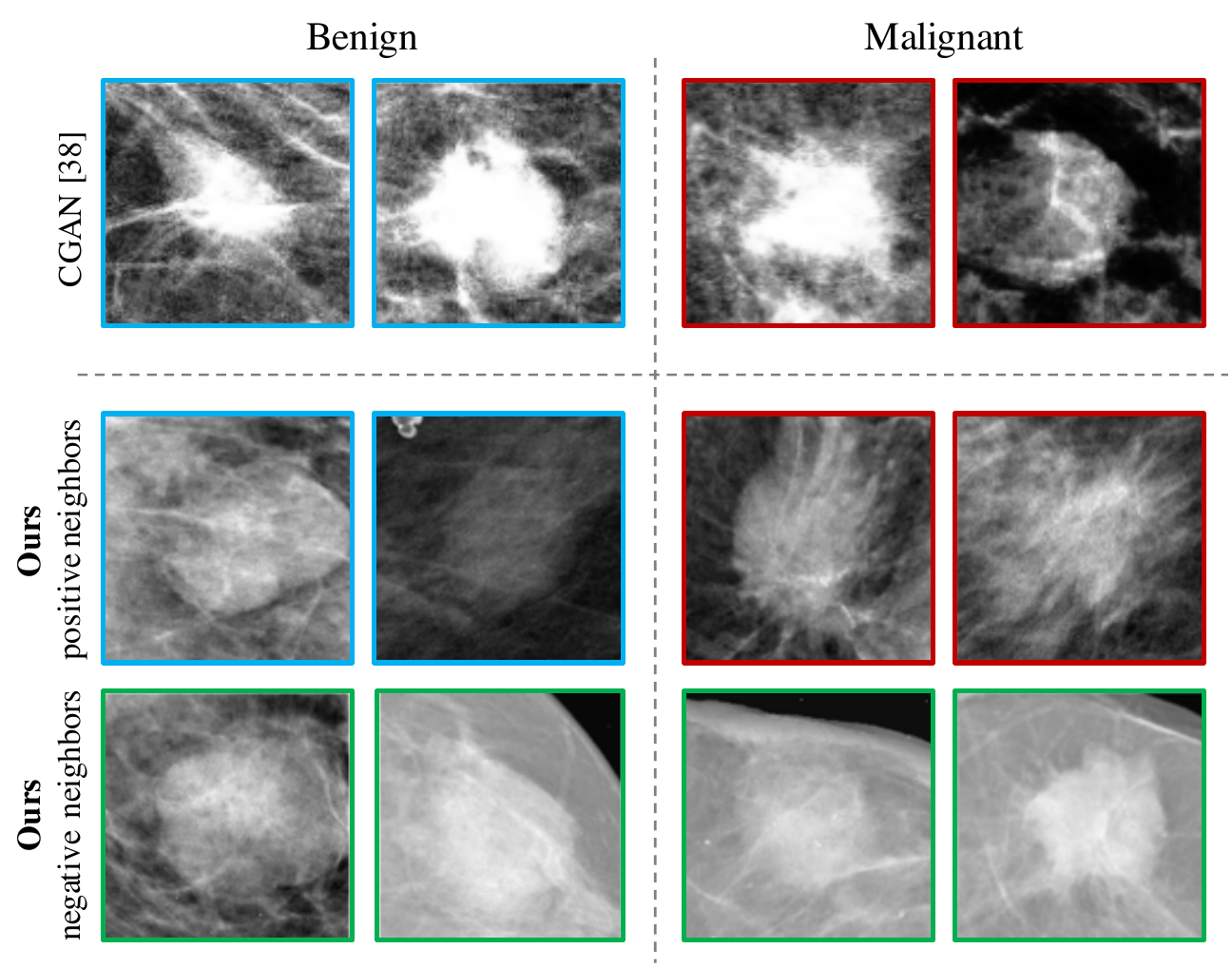}
	\caption{Augmented mass ROIs by Conditional GAN  \cite{isola2017image} (first row), and positive and negative neighbors by our proposed adversarial augmentation method in second and third row respectively. }
	\label{fig:gen_masses}
\end{figure}

The main targets solved by \textsc{COIN} can be formulated as follows:
(1) Given a mass contained mammogram ROI, adversarial augmentation (discussed in Sec. \ref{sec:adversarial_augmentation}) is first employed for each mass category one by one,
so that both on-distribution and off-distribution samples of each class are created: $\mathbf{x}_{i, c} \rightarrow \{\mathbf{x}_{i, c}^+, \mathbf{x}_{i, c}^-\}$, where $\mathbf{x}_{i, c}^+ \in X_c^+$ is positive (indistinguishable from the real masses in $X_c$ by the discriminator) and $\mathbf{x}_{i, c}^- \in X_c^-$ is negative (distinguishable by the discriminator).
(2) For each mass category, with the expanded dataset $\{X_c, X_c^+, X_c^-\}^C$,
the local Signed graph is then constructed.
(3) Based on the results of preceding two steps, contrastive loss is optimized within the localized built signed graph in the deep latent space, learning a nonlinear embedding in the deep latent space $\mathbf{x}_i \rightarrow h(\mathbf{x}_i)$, where manifolds of two categories are maximized. Finally, the latent features are transformed into diagnosis label with a softmax function: $h(\mathbf{x}_i) \rightarrow y_i$.

\subsection{Adversarial Augmentation for Data Scarcity (Q1)}
\label{sec:adversarial_augmentation}

\begin{table}[!t]
\renewcommand{\arraystretch}{1.5}
\caption{BMD performance by constructed deep CNN network with conventional and CGAN \cite{isola2017image} augmentations on benchmark INbreast dataset.}
\centering
\begin{tabular}{l|c|c}
\hline
\bfseries Augmentation Method & \mdseries  Accuracy &\mdseries AUC  \\ 
\hline
Baseline (no augmentation) &  83\% & 0.85   \\
Conventional augmentation  & 87\% & 0.88  \\
CGAN augmentation \cite{isola2017image} & 88\% & 0.89  \\
Proposed augmentation & 89\% & 0.92 \\ 
\hline 
\end{tabular}
\label{tab:cGAN_augmentation}
\end{table}

\subsubsection{Motivation}
As previously mentioned, data scarcity and the high resemblance across benign and cancerous categories of masses are the two major causes \cite{dhungel2016automated} why mammographical CADs are limited, typically with high false positive rates and low sensitivity. 
Recent studies \cite{antoniou2017data, wu2018conditional}, and \cite{shams2018deep} have employed GANs to create new mammogram instances.  In particular, Wu et al. \cite{wu2018conditional} have proposed the use of infilling method, by which generated masses are synthesized in a normal mammogram tissue.  By utilizing class-conditioned GAN, their new samples produced from the generator are forced to be on the same distribution of the original data. 
Yet, they have ignored the importance of surrounding tissues, where textures of blood vessel have imposed a vital role for diagnosing cancerous lesions. This can be the reason of limited improvement over affine augmentation method of their approach.

Thereby, it is natural to directly employ a conditional GAN \cite{isola2017image} to create mass-contained ROIs either from benign or malignant classes, for the purpose of 
enlarging the size of training data and preserving the surrounding contextual features. 
Specifically, the generator in \cite{isola2017image} maps an observed image $\mathbf{x}_{i, c}$ from class $c$ and random noise $\omega$ to the output estimation $\mathbf{x}_{i, c}^+$, i.e. $\{\mathbf{x}_{i, c}, \omega\} \mapsto \mathbf{x}_{i, c}^+$. 
The discriminator involves two mapping components: one is the distinguishing mapping  $\{\mathbf{x}_{i, c}, \mathbf{x}_{i,c}^+\} \mapsto {z}_{i, c}$, where ${z}_{i, c}$ is the predicted probability of being a real data image;
the other is a distance conditional guidance, by which the deep latent features of a created sample is mapped as those of the real data sample, i.e. 
$f(\mathbf{x}_{i, c}^+) \mapsto f(\mathbf{x}_{i,c})$, where $f(\cdot)$ is the non-linear function learned by the CNN. 
As described in \cite{isola2017image}, the generator is constructed with an auto-encoder with skips and the discriminator applies a dual-path CNN architecture with VGG-19 \cite{VGG16and19_2014} as the backbone network \cite{li2016precomputed}.

The generated augmentation samples by the method of conditional GAN \cite{isola2017image} are shown in the first row of Fig. \ref{fig:gen_masses}, and the empirical comparison of classification is shown in Table I.
As shown in Fig. \ref{fig:gen_masses}, the conditional GAN \cite{isola2017image} has performed limited ability in extracting low frequency features, whereas
it focus on the high frequency information when comparing with the original mass samples.
The shape of the augmented masses  are in fact very similar to the realistic ones. 
In addition, the spiculated lines and blood vessels are vividly shown in mass surroundings, and mass boundaries can be seen with high contrast.
Yet, the generated lesions are visually very noisy, especially in the regions within masses, where textual features are merely depicted. As shown in the first row of Fig. \ref{fig:gen_masses}, there is no surrounding tissue have been generated as background  tissue in the last subfigure.  
In order to examine the effectiveness of increasing model discriminativity, we empirically compare the breast mass diagnosis performance (the classification accuracy and the AUC score) in Table \ref{tab:cGAN_augmentation}.   
It can be seen that both augmentation methods have increased the breast mass diagnosis performance over the baseline model by a analogously small margin, nevertheless the model complexity of conditional GAN is much higher than affine transformation. 
  
This limitation by GAN-based methods may stem from neglecting some distinguished samples by the discriminator, which locate very close to the original data distribution. These off-manifold samples are highly similar to the original data, which may confer diverse benefits to classifier discriminativity as being trained along with on-distribution samples.

\subsubsection{Proposed algorithm}
In order to overcome this defect found in previous works and experiments on cGAN \cite{isola2017image}, we desire to enlarge the mammography dataset meanwhile creating more distinctive samples.  
Inspired by Yu et al.'s recent research in solving open-category classification problem \cite{yu2017open},
we propose to use adversarial learning to augment mammographical masses with an optimization free algorithm. 
In this way, we augment the original dataset with both \emph{positive neighbors}, that new instances lie on the original data manifold, \emph{e.g.} $x_5$ and $x_7$ shown in Fig. \ref{fig:diagnet_motivations}; and \emph{negative neighbors}, that augmented samples are off the original data manifold, \emph{e.g.} $x_6$ and $x_8$ in Fig. \ref{fig:diagnet_motivations}. 

Specifically, augmented data samples are generated for each class $c$ separately. For every mass type,  the positive neighbors ${X_c^+}$ and the negative neighbors ${X_c^-}$ are created with the same model but with different objective functions.
Particularly, the positive neighbors ${X_c^+}$ are the generated samples that cannot be separated from $X_c$ by the discriminator, while the negative neighbors ${X_c^-}$ are the ones that can be separated. 
Finally, the expanded dataset for class $c$ is of the form  $\mathcal{X}_c=\{X_c \cup X_c^+ \cup X_c^-\}$, and the whole dataset is $\mathcal{X}=\bigcup_c\mathcal{X}_c$.

In terms of the generator,  the random noise $\omega$ is utilized to corrupt selected seed points, which are a number of randomly selected samples in $X_c$. This step is simply a noise addition, thus no optimization with any objective function is involved.
By applying the generator, new instances, including both the positive neighbors ${X_c^+}$ and the negative neighbors ${X_c^-}$ of samples from class $c$, are created.
All of the new sample nodes are close to the original data points, no matter whether they are positive or negative neighbors.  

After the new instances are generated by the generator, the resulting samples are fed into the discriminator network, which is trained to distinguish the augmented samples and the original data instances.
We adopt a SVM classifier as the discriminator for each type of neighbor of class $c$,
by which the generated samples are discriminated as the ``real'' or ``fake'' category. 
The output of the generator $P_{D}$ ranging with $[0, 1]$ indicates how ``real'' the generated mass is, where $P_{D}=1$ represents real and $P_{D}=0$ denotes generated. 
The corresponding probability score by the SVM is calculated by the logistic sigmoid of the output signed distance, which is formulated as
\begin{equation}
    P_{D}({\mathbf{x}}) =  {\displaystyle\frac {\exp\big({\tilde{d}(\mathbf{x})\big)}}{\exp{\big(\tilde{d}(\mathbf{x})\big)}+1}},
\end{equation}
where $\tilde{d}(\mathbf{x})$ is the signed distance to the decision boundary. 
 
With the built generator and discriminator, we create the new masses one by one, in which
two SVM classifiers for the positive and negative neighbors are trained separately.
Regarding the creation of positive neighbors, let $\mathbf{x}$ be a desired new sample for class 
$c$, and $P_{D}(\mathbf{x}; X_c, X_c^+)$ be the output probability  score of the discriminator trained for positive neighbors. 
At this point, the discriminator aims to generate new samples that are as analogous as possible to the original instances, thereby it is trained on the union of $\mathbf{x}$ and $\{X_c, X_c^+\}$. Note that $X_c^+$ represents the already existing positive neighbors, which is initialized as empty. 
For each training batch, $T$ generated samples $\{\mathbf{x}_t\}_{t=1}^T$ and $T$ original data images $\{\mathbf{X}_c\}_{t=1}^T$ (for the data balanced) are utilized as the input of the discriminator and the weights are updated. After being fully trained, we select only one best generated sample in each batch, according to the objective as follows:
\begin{equation}
\begin{aligned}
\argmax_{\mathbf{x}} P_{D}\big( & \mathbf{x}; X_c, X_c^+\cup \{\mathbf{x}_t\}_{t=1}^T\big) \\ & - \gamma \max \{0, r_1 - \min_{\mathbf{x}_i \in X_c^+} d(\mathbf{x}, \mathbf{x}_i)\}, 
\end{aligned}
\label{eq:positiveneighbors_A}
\end{equation}
where $d(\cdot)$ is a distance measure, and $\gamma$ weights the distance regularization. This regularization term forces the generated points to be different with a minimum distance $r_1$, allowing the generator a better generalization. 

Regarding the creation of negative neighbors,
let $P_{D}(\mathbf{x}, X_c X_c^-)$ corresponds to the output of the discriminator, predicting the possibility of $\mathbf{x}$ labeled as a ``real'' data sample from class $c$.  
$X_c^-$ is the existing negative neighbor set and is initialized as empty. 
In this scenario, the discriminator would like to select the generated samples, which are not only off the original data manifold but also are located close to the original data. In this way, the new samples can provide discriminative information. 
Specifically in a training batch,  we select the desired negative neighbor $\mathbf{x}$ from the $T$ generated samples, according to the objective:

\begin{equation}
\begin{aligned}
    \argmin_{\mathbf{x}} P_{D}\big(&\mathbf{x}; X_c, X_c^-\cup \{\mathbf{x}_t\}_{t=1}^T\big) \\ 
    &+ \gamma \max \{0, r_2 - \min_{\mathbf{x}_j \in X_c^-} d(\mathbf{x}, \mathbf{x}_j)\} \\
&+
 \gamma \max \{0, \min_{\mathbf{x}_i \in X_c} d(\mathbf{x}, \mathbf{x}_i) - r_3\}, 
\end{aligned}
\label{eq:negativeneighbors_A}
\end{equation}
where the distance regularization forces generated points to acquire a minimum distance $r_2$ apart from each other. The added distance restriction forces new points to be scattered close to $X_c$, so that the minimum distance of $\mathbf{x}$ to the original images is at most $r_3$. 
The distance measure $d(\cdot)$ in (\ref{eq:positiveneighbors_A}) and (\ref{eq:negativeneighbors_A}) is set to be the angular cosine distance because of its superior discriminative information \cite{nair2010rectified}.
Let $\rho = \min_{\boldsymbol{x}_i,\boldsymbol{x}_j \in \mathcal{X}_c}d(\boldsymbol{x}_i, \boldsymbol{x}_j)$, then we set the radius parameters $r_1, r_2 = \rho$, and $r_3=3\times \rho$ for $\mathcal{X}_c$. Further $T=200$ and $\gamma$ is $10^{-2}$.

As for the optimization of (\ref{eq:positiveneighbors_A}) and (\ref{eq:negativeneighbors_A}), we employ the derivative free optimization method proposed in \cite{yu2016derivative}, in which the problem of $\argmax_{x\in X}f(x)$ is considered. Instead of calculating the gradients with respect to each parameter, this technique samples a number of solutions of $x$, by which the feedback information is learned for searching for better solutions. The advantage of this method is to optimize problems even with bad mathematical properties, such as non-convexity, non-differentiability and too many local optima \cite{yu2016derivative}.

\subsection{Contrastive Learning to Enhance Discriminativity (Q2)}
\label{sec:contrastive_learning}
Investigators have achieved promising diagnosis performance for mammography by using deep neural networks. Yet one major limiting factor for continued studies is that deep models disregard the structural features of data.   
We consider to integrate the inherent data geometrical factor with CNNs with the merits of contrastive learning. By doing this, samples originated from same distribution are forced to be close whereas samples belonged to different categories are pushed away in the embedding space.  Thus, the model's discriminativity is expected to improve. 

\subsubsection{Motivation}
Contrastive learning was initially proposed to solve the manifold embedding problem in a self-supervised manner \cite{hadsell2006dimensionality} and hence was extensively applied in representation learning \cite{wu2018unsupervised, hjelm2018learning}. This is attributed to its promising performance to improve model's discriminativity through measuring similarities between correlated sample pairs, instead of directly computing sample-wise loss functions (\emph{e.g.} softmax, hinge, or mean squared error loss).
Specifically, for a certain anchor sample, only one positive or negative pair is used for the calculation \cite{he2020momentum}.
Positive pairs can be selected by data augmentation or co-occurence \cite{khosla2020supervised}, while negative pairs are typically data samples uniformly sampled from other classes of data.
Triplet loss \cite{schroff2015facenet} works in similar manner but in a supervised way, where labeled triplets  rather than unlabeled neighboring sample pairs are selected for loss calculation.
Similarly, triplet loss depends on triplet correlated samples, which includes one positive (belonging to the same class with the anchor) and one negative pair (from other classes) \cite{ge2018deep}.
Although contrastive learning is effective to separate dense samples in deep latent space, typical triplet loss is not suitable for classifying mammography breast masses. In fact, random selection of negative and positive pairs can lead to worse generalization over the baseline, as the margin of mammogram manifolds across different classes are very close. 
On the contrary, with the use of manifold learning approximated by a designed local Signed graph, contrastive learning is able to preserve manifold locality knowledge, thus maximizing the manifold margin through the penalty involved by the selected neighboring positive and negative samples.

\subsubsection{Signed Similarity Graph}
Graph embedding is trained with distributional context knowledge, which can boost performance in various pattern recognition tasks. Here, we aim to incorporate the signed graph Laplacian regularizer \cite{chen2018learning} to learn a discriminative datum representation $\mathcal{H}(\mathcal{X})$ by a deep neural network, where discriminative here means that the intra-class data manifold structure is preserved in the latent space and the inter-manifold (slightly different) margins are maximized.

Using the supervision of the adversarial augmentation in section \ref{sec:adversarial_augmentation}, we build a Signed graph upon the expanded data $\mathcal{X}$.
Given $\mathcal{X}_c = \{X_c, X_c^+, X_c^-\}$ for class $c$,  and all other classes data
$\mathcal{X}_{-c} = \bigcup_{t=1,\cdots,C\\; t\neq c}\{X_t, X_t^+, X_t^-\}$, 
for $\forall \boldsymbol{{x}}_i \in \mathcal{X}_c$, the corresponding elements in the Signed graph is built as follows:
\begin{equation}
\label{eq:diagnet_adjacency}
\phi_{ij}=\begin{cases}
+1,  & \boldsymbol{{x}}_j \in \{X_c \cup X_c^+\}_i^{n^+}, \\
-1,  & \boldsymbol{{x}}_j \in \{X_{-c} \cup \mathcal{X}_c^-\}_i^{n^-}, 
\end{cases}
\end{equation} 
where the $\{\cdot\}_i^{n^+}$($\{\cdot\}_i^{n^-}$) denotes the corresponding $n^+$ ($n^-$) nearest neighborhood of $x_i$ to approximate the locality of the manifold.

\subsubsection{Triplet contrastive loss}
\begin{figure}[!t]
    \centering
    \includegraphics[width=0.49\textwidth]{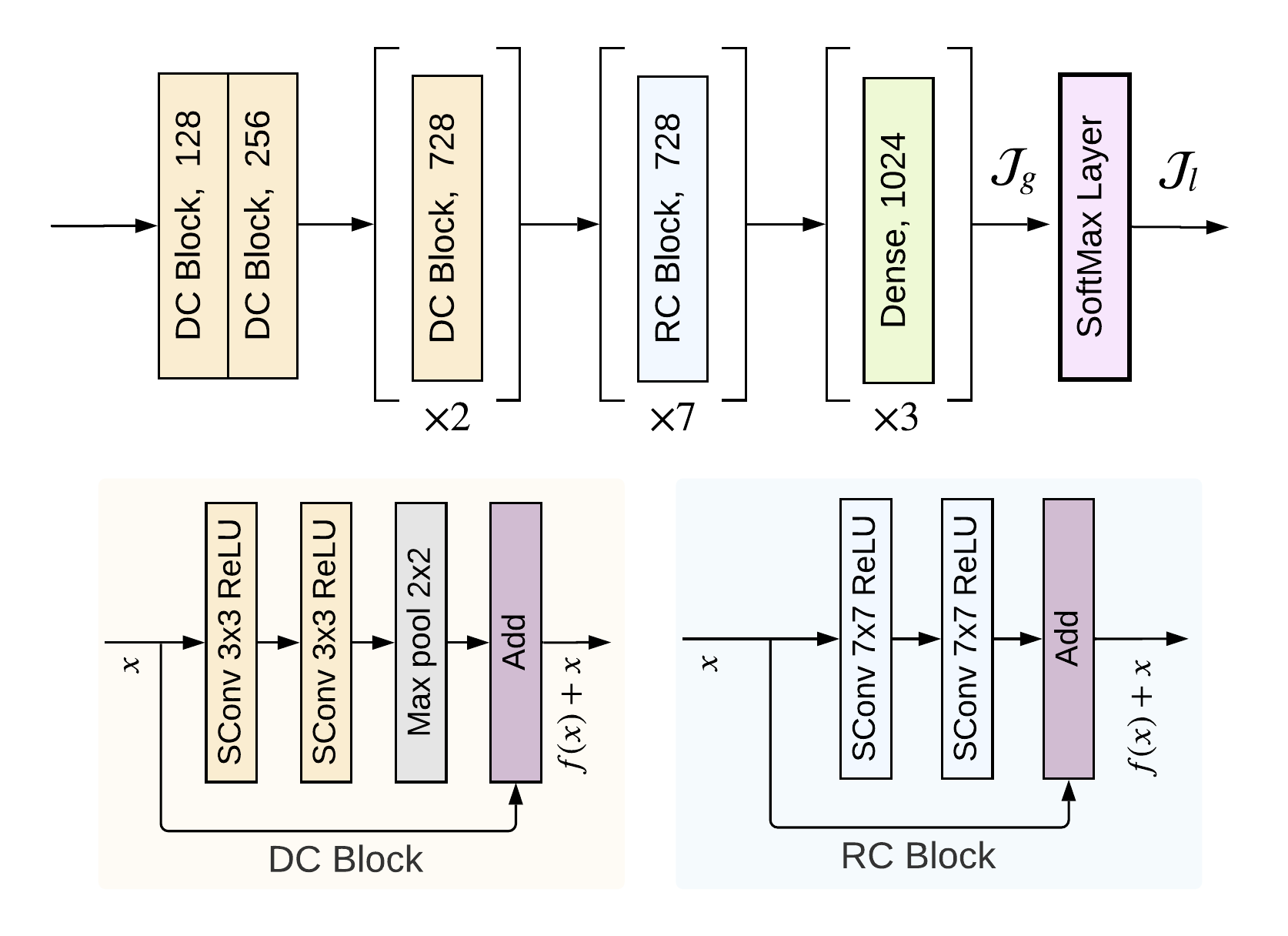}
    \caption{The deep neural network architecture constructed in COIN to extract deep latent features.  ``DC block" represents a down-sampling convolutional block, ``RC block" is a residual convolutional block, and ``SConv" is separable convolutions.}
    \label{fig:diagnet_cnn}
\end{figure}

Then, we compute the structure preservation in the deep representation space (directly behind the softmax layer as shown in Fig. \ref{fig:diagnet_cnn}) $\mathcal{H}=\{{h}(\boldsymbol{x}_i)\}_{i=1}^N$, where $N = |\mathcal{X}|$. The Signed graph Laplacian regularizer is defined as following:
\begin{equation}\label{eq:diagnet_graph}
\mathcal{J}_g(\mathcal{X}, \Phi) = 
\sum\limits_{i,j} \begin{cases}
\phi_{ij} \cdot dist({h}(\boldsymbol{x}_i), {h}(\boldsymbol{x}_j)), 
\text{    if } \phi_{ij} > 0
\\
\begin{aligned}
\max \big(0, m + \phi_{ij} \cdot dist({h}(&\boldsymbol{x}_i), {h}(\boldsymbol{x}_j)) \big), \\ &\text{if } \phi_{ij} < 0,
\end{aligned}
\end{cases}
\end{equation}
where $dist(\cdot)$ is a distance metric for the dissimilarity between ${h}(\boldsymbol{x}_i)$ and ${h}(\boldsymbol{x}_j)$. It encourages similar examples to be close, and those that are dissimilar to have a distance of at least $m$ to each other, where $m>0$ is a margin.

Note that instead of calculating the manifold embedding by solving an eigenvalue decomposition, we learn the embedding $\mathcal{H}$ by a deep neural network. Specifically, inspired by the depth-wise separable convolutions \cite{chollet2017xception} that are extensively employed to learn mappings with a series of factoring filters, we build stacks of depth-wise separable convolutions with similar topological architecture to that in \cite{chollet2017xception} to learn such deep representations (Fig. \ref{fig:diagnet_cnn}).

Therefore, by minimizing (\ref{eq:diagnet_graph}), it is expected that if two connected nodes $\boldsymbol{{x}}_i$ and $\boldsymbol{{x}}_j$ are from the same class (i.e. $\phi_{ij}$ is positive), ${h}(\boldsymbol{x}_i)$ and ${h}(\boldsymbol{x}_j)$ are also close to each other, and vice versa. Benefiting from such learned discriminativity, we train a simple softmax classifier to predict the class label, i.e.,
\begin{equation}\label{eqs:label_crossentropy}
\mathcal{J}_l = -\frac{1}{N}\sum_{i=1}^N\sum_{c=1}^C
{\delta_c(y_i)\log P\big(y_i\mid \boldsymbol{{x}}_i;\boldsymbol{\theta} \big)},
\end{equation}
where $\delta_c(y_i)=1$ when $y_i=c$, and $0$ otherwise; $\boldsymbol{\theta}$ is the parameter set of the neural network. 

\subsubsection{Total Loss}
Finally, by incorporating the Signed Laplacian regularizer (\ref{eq:diagnet_graph}) and  the classification loss  (\ref{eqs:label_crossentropy}), the total objective of \textsc{DiagNet} is accordingly defined as:
\begin{equation}\label{eq:signedlap_loss}
\mathcal{J} =   \mathcal{J}_l + \lambda\mathcal{J}_g,
\end{equation}
where $\lambda \geq 0$ is the regularization trade-off parameter which controls the smoothness of hidden representations.

\subsection{Network Architecture and Implementation}

The proposed CNN model is constructed with the architecture shown in Fig. \ref{fig:diagnet_cnn}. 
In the first four convolutional layers, down-sampling convolutional blocks (DC blocks) involve two separable convolutions are employed.
Specifically, the separable convolution operators decompose $3\times3$ convolutions into consecutive $3\times1$ and $1\times3$ operations. After that a pooling layer halves the spatial size of the feature maps. The output of the down-sampling layer is then obtained by the transformation of the ReLU nonlinearity.  The four DC blocks altered the original input $224\times224\times1$ into feature maps with spatial sizes $112\times112\times128$, $56\times56\times256$, $28\times28\times728$, and $14\times14\times728$ respectively. 
Sequentially, seven separable convolutional layers are padded, reducing the total number of parameters, before three fully connected layers with the numbers of neurons are all $1024$. The obtained latent features of the enlarged dataset are then regularized with the proposed contrastive loss in Sec. \ref{sec:contrastive_learning}.  
Finally, the learned features are classified into binary classes ($0$ denotes "Benign`` and $1$ represents "Malignant``).

\begin{figure*}[!t]
	\normalsize
	\centering
	\subfloat[Configurations of $(n^+, n^-)$]
    {
        \includegraphics[width=0.38\textwidth]{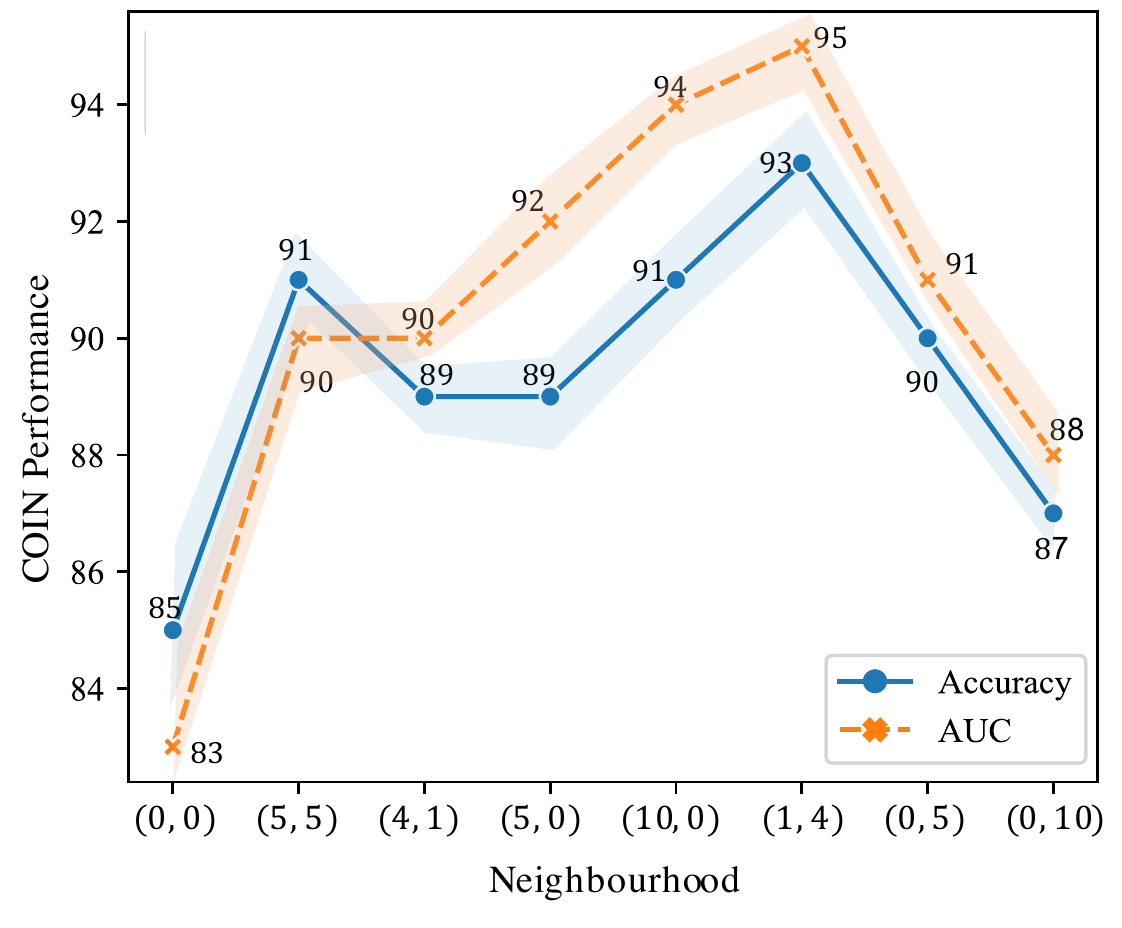}
        \label{fig:compare_n}
    } 
    \hfil
    \subfloat[Configurations of $\lambda$]
    {
        \includegraphics[width=0.38\textwidth]{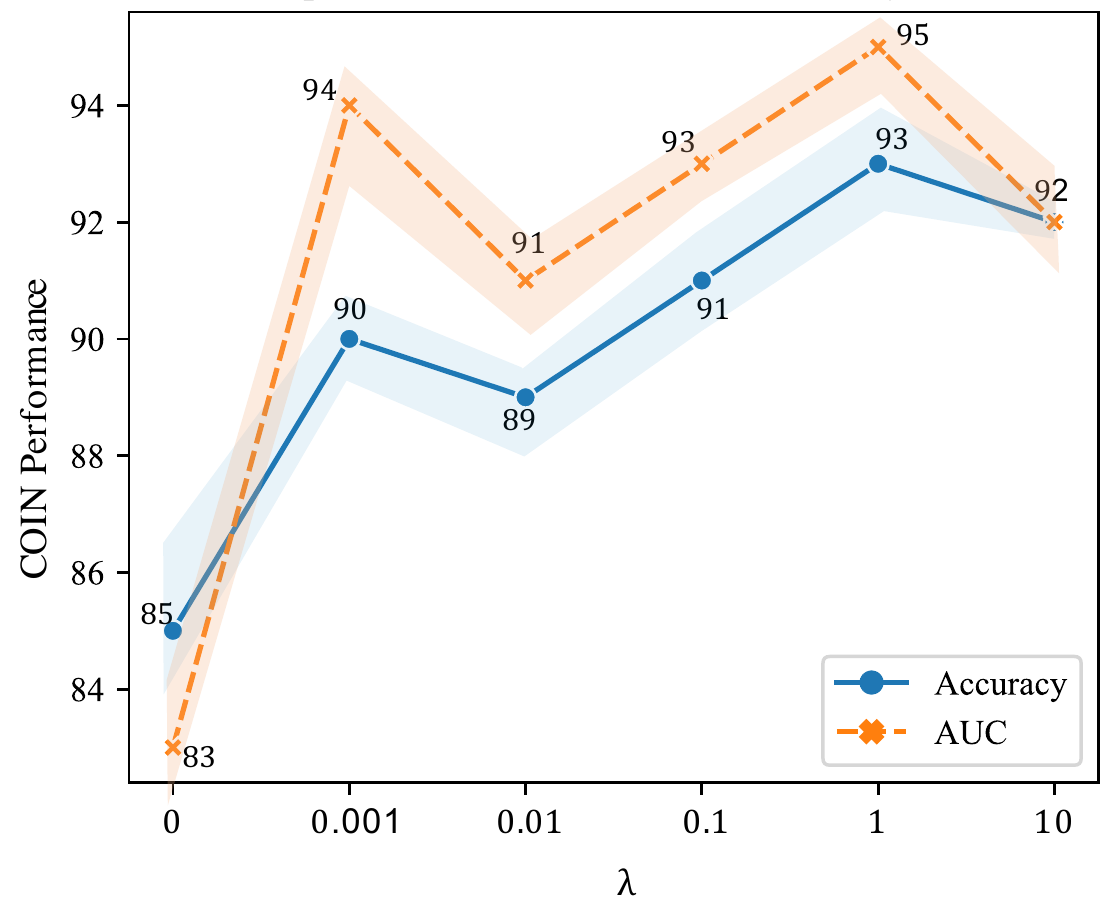}
        \label{fig:compare_lambda}
    }
\caption{BMD Performance (accuracy and AUC score) of \textsc{COIN} on INBreast v.s. various hyper-parameters $\lambda$, $n^+$ and $n^-$. (a) shows the performance with different $n^+$ positive neighbors and $n^-$ negative neighbors when $\lambda$ equals 1, and (b) depicts various regularizer parameter $\lambda$ with $n^+=1$ and $n^-=4$.}
\label{fig:diagnet_config}
\end{figure*}

\begin{figure*}[!t]
	\normalsize
	\centering
    \subfloat[COIN ($n^+=0$, $n^-=0$)]
    {\includegraphics[width=0.31\textwidth]{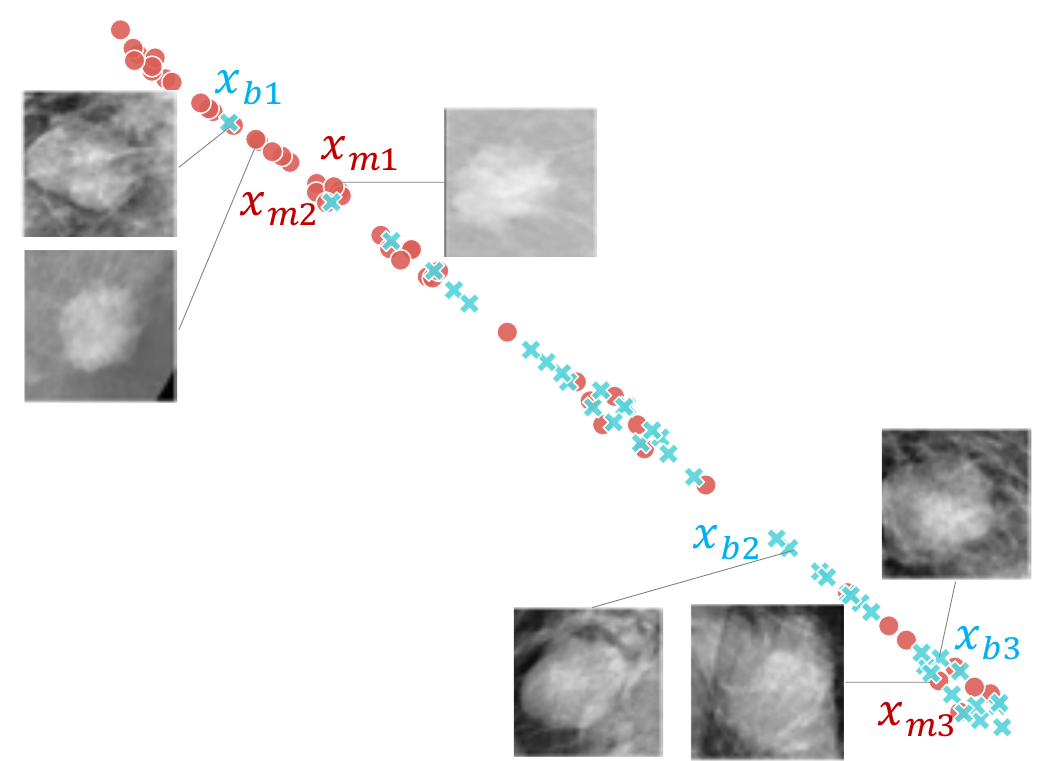}\label{fig:embedtest_cnn}}
    \hfill
    \subfloat[COIN ($n^+\neq0$, $n^-=0$)]
    {\includegraphics[width=0.31\textwidth]{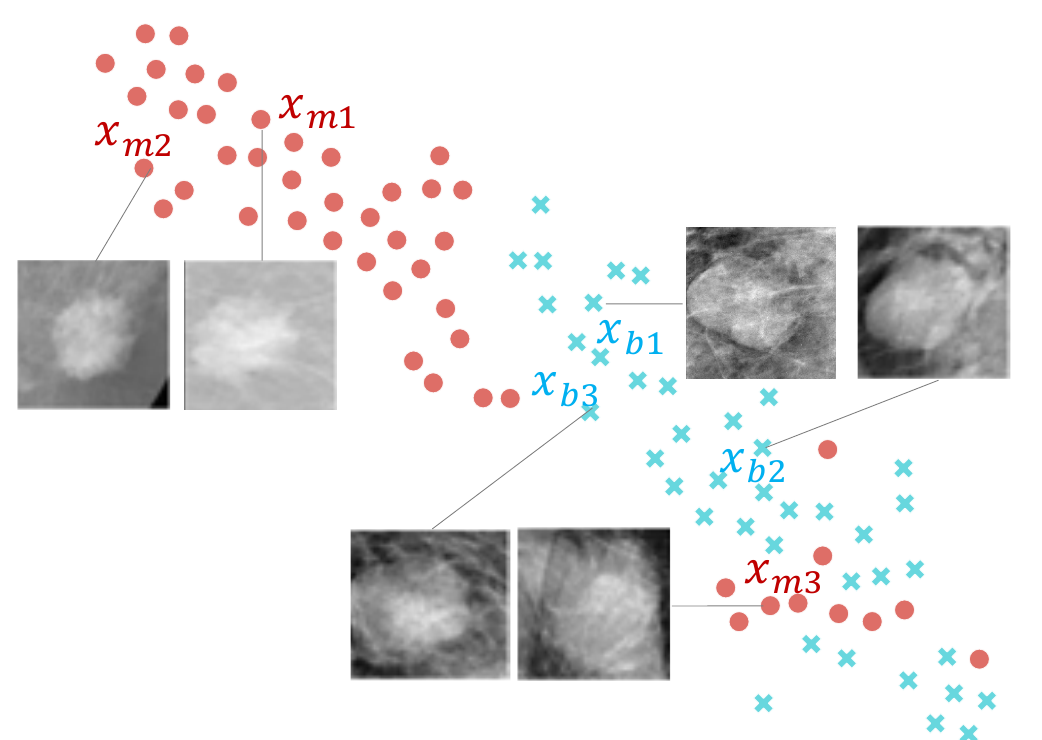}\label{fig:embedtest_coinpositive}}
    \hfill
    \subfloat[COIN ($n^+\neq0$, $n^-\neq0$)]
    {\includegraphics[width=0.31\textwidth]{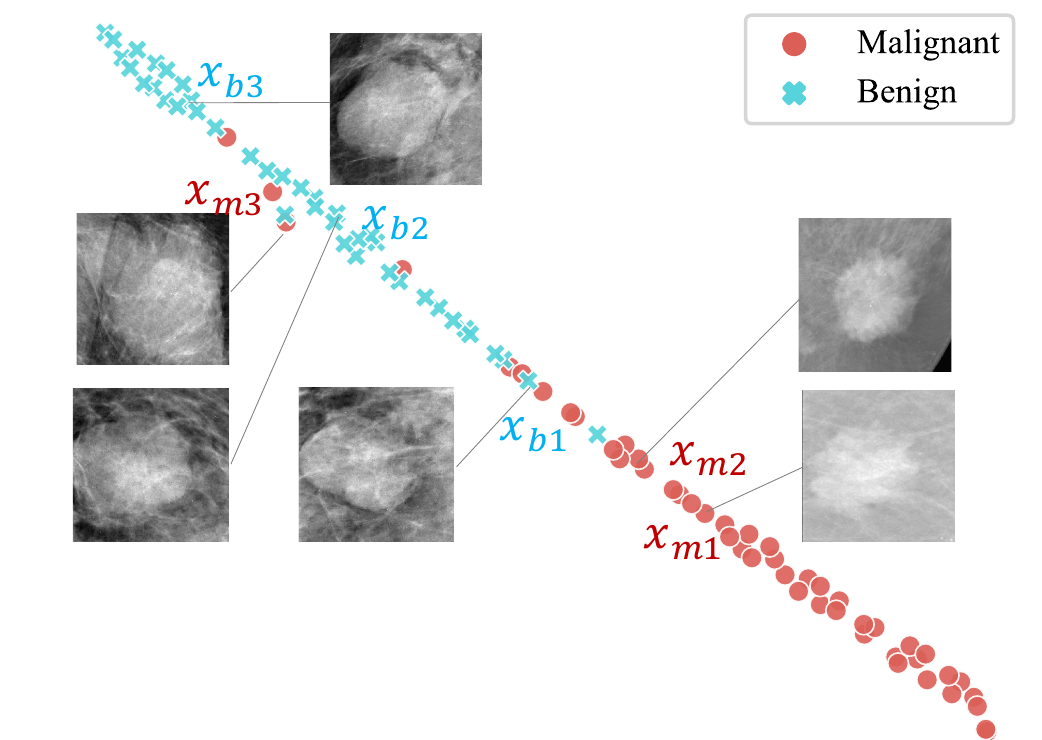}\label{fig:embedtest_coin}}
\caption{t-SNE plots for the test set of INbreast dataset. (a), (b) and (c) show the embbedings of latent features trained by COIN with various learning configurations.}
\label{fig:embed_testdata}
\end{figure*}

\section{Experiments}
In this section, extensive experiments will be implemented to validate the proposed algorithm. We first examine the quality of generated masses  from both adversarial augmentation modules. 
We then expand the original dataset with the augmented data, and build the Signed graph. To better evaluate the performance, we validate the proposed algorithm on the small FFDM mammography dataset: the INbreast dataset \cite{moreira2012inbreast}.

\subsection{Adversarial Augmentation Performance}

To visually examine the quality of generated images by the proposed adversarial augmentation strategy, Fig. \ref{fig:gen_masses} show the augmented examples for benign and malignant categories (blue stands for benign and red represents malignant masses). 
It is noticeable that the difference between positive (second row) and negative neighbors (third row) within each category is subtle. Visually, it is very difficult to differentiate them within each mass type, not only with the masses themselves but also with the contextual or background tissues. This indicates that the generated negative neighbors are challenging to recognize, thus they tend to play an important role in increasing model's discriminative ability.  
When we compare the generated samples by our proposed method  with cGAN generated samples (first row), we can notice that the generated positive and negative samples of both benign and malignant categories are less noisy with more balanced concentration on low and high frequency signals.
When observing the left column subfigures, it can be noticed that, both negative and positive neighbors of benign masses are in oval or round shape with relatively smooth boundaries, which are very similar to that of original INbreast data (Fig. \ref{fig:embed_originaldata}). Additionally, the textual and contextual features of generated and realistic samples are visually highly alike. 
From the right column in Fig. \ref{fig:gen_masses}, it can be seen that the shape of our resulting malignant masses (including both positive and negative neighbors) are mostly irregular, and the boundaries are fuzzy with spiculated vessels. These characteristics are identical to malignant masses in original INbreast dataset (Fig. \ref{fig:embed_originaldata}). 

In order to further evaluate the effectiveness of the proposed Adversarial Augmentation, we design a series of experiments to test the discriminativity of generated mass samples. As shown in Tab. \ref{tab:cGAN_augmentation}, we evaluate the classification performance with different augmentation algorithms in the proposed CNN architecture (Fig. \ref{fig:diagnet_cnn}), which include original INbreast data (baseline), conventional augmentation (flips and rotations), CGAN augmentation \cite{isola2017image} and the proposed adversarial augmentation (positive neighbors only, i.e. $(n^+, n^-)$ is $(5, 0)$ and $\lambda = 0$). Note that we optimize the CNN model with cross-entropy loss. From the Tab. \ref{tab:cGAN_augmentation}, we can notice that all augmentation algorithms have improved the classification performance when comparing with the baseline model. The conventional augmentation and CGAN \cite{isola2017image} have achieved similar discriminative performance, whereas the proposed augmentation has outperformed other listed methods in both accuracy rate and AUC score. The proposed adversarial augmentation algorithm has achieved 89\% accuracy and 0.92 AUC score.

\subsection{Signed Graph Laplacian performance} 
Determining the optimal values of hyper-parameters is a big challenge in deep learning. To explore \textsc{COIN}'s performance with different Signed graph configurations, the values of the number of positive neighbors $n^+$ and the number of negative neighbors $n^-$ are first grid searched with fixed regularization parameter $\lambda=1$, as shown in Fig. \ref{fig:compare_n}. 
The best performance occurs when $n^+=1$ and $n^-=4$, which increases at least by 8\% in the accuracy rate and by 12\% in the AUC score when compared to no graph regularization. 
This confirms the effectiveness of using the signed graph regularization and also validates the importance of negative neighbors to improve the discriminativity and maximize the manifold margin. 
In addition, results show that the \textsc{DiagNet} achieves good performance only when both $n^+$ and $n^-$ are considered in the corresponding singed graph construction. 
Furthermore, we fix the best performing Signed graph configuration to evaluate the $\lambda$ value and obtain the best AUC score and accuracy at $\lambda=1$.
These results indicate that the deep latent features extracted by the deep network and the data inherent structural features are both important when diagnosing the malignant breast masses from the benign ones. 

\begin{table}[!t]
\renewcommand{\arraystretch}{1.2}
\caption{Breast Mass Diagnosis performance comparisons of the proposed \textsc{DiagNet} and relative state-of-the art methods on INbreast test set.}
\label{table_results}
\centering
\begin{tabular}{l|c|c}
\hline
\bfseries Methodology  &\mdseries Accuracy& \mdseries AUC  \\ 
\hline
Domingues \textit{et al.} (2012) \cite{domingues2012inbreast} & 89\% & N/A   \\
Dhungel \textit{et al.} (2016) \cite{dhungel2016automated}    & 91\% & 0.76  \\
Zhu \textit{et al.} (2017) \cite{zhu2017deep}                 & 90\% & 0.89  \\
Shams \textit{et al.} (2018) \cite{shams2018deep}       & 93\% & 0.92 \\
Li \textit{et al.} (2019)  \cite{li2019deep}    & 88\% & 0.92   \\
\hline 
\textsc{COIN} (ours)   & $\boldsymbol{93.4}\pm1.9\%$ & $\boldsymbol{0.950}\pm0.02$ \\ 
\hline 
\end{tabular}
\end{table}   

To visually observe the performance of data manifold learning, we further explore the learned features embedding plotted by t-SNE for test set (Fig. \ref{fig:embed_testdata}). 
For the purpose of ablation study, we explore the performance of COIN with different learning configurations. For instance, Fig. \ref{fig:embedtest_cnn} shows COIN without any intra class or inter class Signed graph regularization provided by positive or negative neighbors, respectively. 
Fig. \ref{fig:embedtest_coinpositive} shows the learning performance when COIN is only regularized by intra class regularization, \emph{i.e.} without the usage of negative neighbors. 
And Fig. \ref{fig:embedtest_coin} illustrates the COIN learning when both intra and inter class regularization are employed. 
When compare these three conditions,  the worst performance is obtained when there is no regularization (Fig. \ref{fig:embedtest_cnn}), by which samples of two categories are highly intersected.  When the model is trained with intra class regularization (Fig. \ref{fig:embedtest_coinpositive}), it achieves a better discminativity performance, in which 15\% samples are mis-classified. COIN with both negative and positive regularization (Fig. \ref{fig:embedtest_coin}) has achieved the best embedding of the test data, where 82 out of 88 masses or approximately 93\% test samples are correctly identified.  
Additionally, we have attached the original mass examples for some randomly selected misclassified masses in Fig. \ref{fig:embed_testdata}. We can notice that, the misclassified malignant mass sample by COIN are particularly similar to those benign masses surrounding it, and vice versa. This indicates that COIN can correctly categorize breast masses in most cases, apart from extremely hard example.  

\subsection{Comparison to the state-of-the-art} 
Finally, to further explore the effectiveness of COIN,  
we compare the proposed algorithm with the state-of-the-art methods in Tab. \ref{table_results}, where results of other works are taken from their original papers. It shows that,  \textsc{COIN} has outperformed the state-of-the-art with mean accuracy 93.4\% and AUC score 0.95. When compared with the second best algorithm  \cite{shams2018deep},  \textsc{COIN}'s AUC score is significantly higher (3\%) with experiments on the whole dataset without any pre-processing, post-processing or transfer learning.

\section{Conclusions}
In this paper, 
we have proposed a novel deep framework \textsc{COIN} to address the two crucial challenges of BMD problem, \emph{i.e.} data scarcity and data entanglement.
\textsc{COIN} integrates adversarial augmentation and contrastive learning.
In particular, the proposed adversarial augmentation dose not only enlarge the dataset, but also enhances the discriminativity for the diagnosis model. 
The proposed contrastive learning merits the model's distinguishable ability further via exploiting the manifold geometry of data, which is valuable for mammography lesions of high resemblance.   
Experiments have shown that \textsc{COIN} surpasses the state-of-the-art algorithms for BMD problem.

\bibliographystyle{IEEEtran}
\bibliography{reference/reference}
\end{document}